\documentclass[11pt]{article}
\usepackage{coling2020}
\usepackage{times}
\usepackage{url}
\usepackage{latexsym}
\usepackage{multirow}
\usepackage{graphicx}
\usepackage{amsmath}

\usepackage[ruled,vlined]{algorithm2e}
\usepackage{microtype}
\SetKwInput{KwInput}{Input}
\SetKwInput{KwOutput}{Output}
\colingfinalcopy

\title{To What Degree Can Language Borders Be Blurred \protect\\  In BERT-based Multilingual Spoken Language Understanding?}

\author{Quynh Do, Judith Gaspers, Tobias Röding, Melanie Bradford\\
  Amazon Alexa\\
  {\tt \{doquynh,gaspers,rodingtr,neunerm\}@amazon.com} \\
}

\date{}

\begin{document}
\blfootnote{
    %
    %
    %
    %
    %
    %
     \hspace{-0.65cm}  
     This work is licensed under a Creative Commons 
     Attribution 4.0 International License.
     License details:
     \url{http://creativecommons.org/licenses/by/4.0/}.
}

\maketitle

\begin{abstract}
This paper addresses the question as to what degree a BERT-based multilingual Spoken Language Understanding (SLU) model can transfer knowledge across languages. Through experiments we will show that, although it works substantially well even on distant language groups, there is still a gap to the ideal multilingual performance. In addition, we propose a novel BERT-based adversarial model architecture to learn language-shared and language-specific representations for multilingual SLU. Our experimental results prove that the proposed model is capable of narrowing the gap to the ideal multilingual performance.
\end{abstract}


\section{Introduction}
Recently, modern voice-controlled devices with virtual assistants like Alexa, Siri, and Google Assistant, have been working their way into every activity of our daily life from playing a song to driving a car. 
Spoken Language Understanding (SLU) models interpreting the semantic meaning conveyed by an user's spoken utterance, are the AI brains of these devices.
The task of SLU is usually divided into two sub-tasks, namely intent classification (IC), which identifies the intent of an user's utterance; and slot filling (SF), which extracts semantic constituents from the utterance. Consider an annotated utterance from the ATIS dataset \cite{atis}: 
\begin{quote}
    \textbf{city}~~~~~~ [$_{\bf{O}}$~where] [$_{\bf{O}}$~is] [$_{\bf{B-airport\_code}}$~MCO]
\end{quote}
 The SF sub-task should classify ``where'' and ``is'' as ${\bf{O}}$ (i.e. Other); and ``MCO'' as ${\bf{B-airport\_code}}$, while the IC sub-task should identify \textbf{city} as the intent. Current state-of-the-art SLU models are mostly DNN-based joint models of the two sub-tasks \cite{Liu2016AttentionBasedRN,do19,chen19}. These models usually contain two individual decoders to detect slot and intent labels on top of a shared encoder. As in many other NLP fields, the BERT encoder-decoder architecture has shown its strong ability in capturing contextual information from data sources to improve SLU performance \cite{chen19}. For common SLU use cases, the BERT encoder is pre-trained on a huge amount of unlabeled texts and then fine-tuned on SLU training data.
 
Since the voice-controlled device market has been expanding at an incredible rate all over the world, there is a rising need of fast language expansion for SLU models. Multilingual technology, which allows the development of a single model for multiple languages, and transferring knowledge from a data resource to languages other than its own, is currently one of the best solutions to meet the increased need.
 Traditionally, an SLU model is trained on supervised monolingual data leading to the fact that a virtual assistant often can use only one language within a working session. In contrast, a multilingual SLU model trained on supervised multilingual data, can provide the virtual assistant the ability to talk in multiple languages within a working session. Moreover, from a development perspective, multilingual modeling helps not only to reduce the number of models to build, but also to reduce supervised data needs by transferring knowledge across languages. However, as languages differ from each other in various aspects from lexicon to syntax, the cross-language knowledge transfer of the current multilingual techniques may still be limited. This raises the questions of: i) what an ideal cross-language knowledge transfer would be; ii) what a naive cross-language knowledge transfer would be; and iii) to what degree a multilingual model can transfer knowledge across languages.


As one of the most successful multilingual techniques recently, multilingual BERT (mBERT) \cite{mBERT} is being used more and more commonly in natural language processing models in both zero-shot\footnote{The model is trained in one language and tested on another.} and multilingual\footnote{The model is trained on multiple languages.} scenarios \cite{wu-dredze-2019-beto,multiatis++}. Since mBERT is trained without any cross-lingual objectives and does not leverage aligned data, its strong cross-lingual abilities are surprising and have, in turn, spurred research aiming to understand why it is able to achieve them 
 \cite{pires2019multilingual}. In this paper, we, however, focus on addressing the question of to what degree it can blur the language borders in the \textit{multilingual} SLU setting by comparing its ability to transfer knowledge across languages with the ideal and naive cases. In addition, as mBERT is pre-trained for general purpose, we are also interested in how to improve its multilinguality for the particular use case of multilingual SLU.


Our contribution in this paper is two-fold: \textit{First}, by performing a wide range of experiments with bilingual and trilingual SLU models, we show that although mBERT is substantially good in blurring language borders even on distant language groups, there is still a gap to the ideal multilingual performance. \textit{Second}, we propose a novel adversarial model architecture to learn language-shared and language-specific representations on top of mBERT representations when fine-tuning on SLU data. The experimental results prove that our proposed approach can 
narrow the gap to the ideal multilingual performance.

\section{Related Work}\label{sota}
As one of the most successful model architectures in natural language understanding recently, BERT models have been explored  in various researches for SLU \cite{chen19,multiatis++,gaspers2020data}. In these works, SLU is considered as a downstream task of a trained BERT model. In particular, after being pre-trained on a large amount of unlabelled data, the BERT embeddings and encoders are fine-tuned on supervised SLU data together with two SLU-adapted decoders for the IC and SF sub-tasks.  


Due to the fast expansion of voice-controlled device market, there has been a rising interest in cross-lingual and multilingual SLU modeling. In the former direction, supervised data from one or multiple source languages is leveraged to improve the SLU performance on a target language \cite{johnson-etal-2019-cross,do-gaspers-2019-cross,multiatis++}. In the later direction, SLU models are trained on supervised multilingual data for multiple target languages \cite{multiatis++}. To obtain a multilingual model, \newcite{multiatis++} simply initialized the embeddings and encoder of a BERT-based SLU model by mBERT's parameters, and then trained the full model on a mix of supervised datasets from multiple languages. Noticeably, the SLU model used in \newcite{multiatis++} was designed towards cross-lingual scenarios using a soft-alignment method to improve the slot projection between source and target languages. In this paper, we focus on only multilingual SLU modeling. Unlike \newcite{multiatis++}, our adversarial model is designed specially for multilingual model building.





The impressive abilities of mBERT in cross-lingual and multilingual natural language understanding applications, have recently led to an increasing body of research analyzing how it is able to achieve those. For instance, \newcite{pires2019multilingual} probed the cross-linguality of mBERT using zero-shot transfer learning on morphological and syntactic tasks and found that mBERT is able to create multilingual representations. However, to the best of our knowledge, probing multilingual representation learning for downstream tasks, like SLU, and addressing the question of how far it is from the ideal expectation,  have not yet been explored in literature. 

Adversarial approaches aiming to account for linguistic differences across languages by dividing the model into language-shared and language-specific representations, have been explored for the SLU sub-tasks. Recently, \newcite{8990095} investigated the sub-tasks in isolation using BiLSTMs and focused on improving SLU for low-resource languages. Meanwhile, \newcite{chen-etal-2019-multi-source} explored BiLSTMs to improve named entity recognition which is close to the SF sub-task of SLU. In this work, instead of isolating the two sub-tasks, we propose a joint model based on mBERT for multilingual use cases. 


\section{BERT-based multilingual SLU models}
\label{lbl:method}
In this section, we describe the two multilingual SLU models evaluated in this paper. 

\subsection{SLU as a downstream task of BERT}
\label{slu-bert}
\begin{figure}[h]
\centering
\includegraphics[width=10cm]{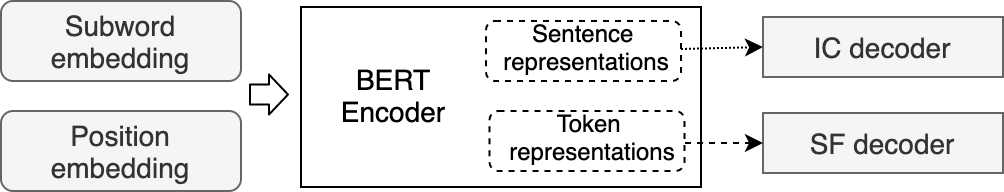}
\caption{A standard BERT-based SLU model architecture.}
\label{fig:bert_slu}
\end{figure}
Fig. \ref{fig:bert_slu} shows our BERT-based SLU model architecture. 
Like \newcite{chen19} and \newcite{gaspers2020data}, our model consists of a BERT encoder receiving sub-word and position embedding layers as inputs, an IC decoder to identify intent labels, and an SF decoder to predict slot labels. In particular, given an utterance $u$, the final hidden states of the tokens $h_1\ldots h_T$ produced by the encoder, are used as the token representations. The sentence presentation, denoted by ${s_u}$, is computed from $h_1\ldots h_T$ by using max-pooling.

${s_u}$ is passed through the IC decoder, which is composed from  a feed-forward of 2 dense layers ($\text{FFN}^I$) and a linear output layer to predict the intent:
\begin{equation}
\begin{split}
\hat{s}_u  =  \text{FFN}^I(s_u) \\
p^i(.|u)  =  \text{softmax}({\bf W}^I\hat{s}_u + b^I)
\end{split}
\label{eq:pri}
\end{equation}

$h_1\ldots h_T$ are passed through the SF decoder, which is composed from a position-wise feed-forward of 2 dense layers ($\text{FFN}^S$) and a CRF layer on top \cite{zhou-xu-2015-end} to predict slot labels:
\begin{equation}
\begin{split}
\hat{h}^t  =  \text{FFN}^S(h^t) \\
\end{split}
\label{eq:pri}
\end{equation}
The CRF layer takes $\hat{h}^t$ as inputs, and estimates a transition matrix modeling the dependence among adjacent labels. The slot labels are predicted by the traditional Viterbi decoding. 

The model is trained by optimizing the joint loss $L  =  L_i + L_s$ where $L_i$ and $L_s$ are the cross entropy loss for intent identification and the CRF loss for slot classification, respectively.

As a downstream task of BERT, the embedding and encoder layers are initialized by a pre-trained BERT language model. The full model is  trained on a supervised SLU dataset.



\subsection{A simple approach for multilingual modeling}
\label{lbl:method:standard}
Thanks to the strong multilingual abilities of mBERT, we simply use mBERT for the model described in \ref{slu-bert} to obtain a simple multilingual SLU model. In particular, the embedding and encoder layers are initialized by the parameters of a pre-trained mBERT. The full model is then trained on a mix of supervised data from multiple languages.



\subsection{An adversarial approach for multilingual modeling}
\label{lbl:adv}
\begin{figure}[h]
\centering
\includegraphics[width=15cm]{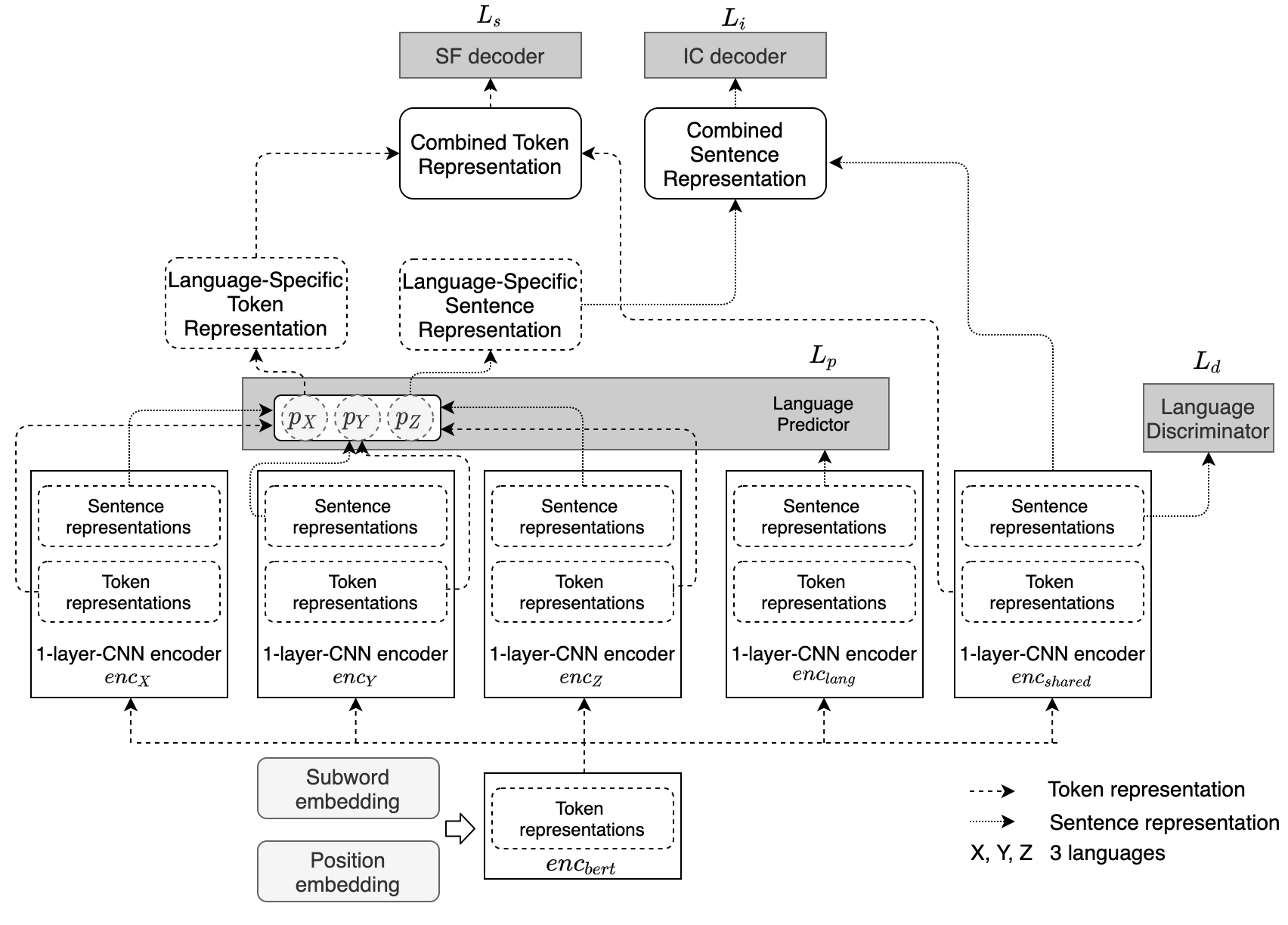}
\caption{An adversarial model architecture for BERT-based trilingual SLU.}
\label{fig:adv_bert_slu}
\end{figure}

The simple multilingual model, as described in Sec. \ref{lbl:method:standard}, learns language-shared token and sentence representations from multilingual training data via its BERT encoder. However, as languages differ from each other in various aspects, using only language-shared representations may be \textit{not} enough to reach the optimal performance in all of the target languages. Our hypothesis is that encouraging a model to learn \textit{both language-shared and language-specific representations} will narrow the gap to the ideal multilingual SLU model in which knowledge can be transferred freely across language borders.

Motivated by the success of adversarial approaches in natural language processing, we propose a novel BERT-based  adversarial SLU model using a single BERT encoder to learn language-shared representations across all target languages, and  multiple CNN encoders to learn language-specific representations for each of the target languages. The language-shared and language-specific representations are concatenated before being fed to the IC and SF decoders. The full model is trained via an adversarial training strategy. Fig. \ref{fig:adv_bert_slu} show our model architecture for a trilingual use case.

In a general use case targeting $N$ languages $l_1, \ldots , l_N$, the model has $N+3$ encoders:  First, it contains a BERT encoder denoted by $enc_{bert}$, which can be initialized by pre-trained BERT parameters. 
Second, we use $N$ 1-layer CNN encoders to learn language-specific representations for $l_1, \ldots , l_N$, denoted by $enc_{l_1} , \ldots , enc_{l_N}$, respectively. Third, an 1-layer CNN encoder denoted by $enc_{lang}$ is used to learn features for predicting the language of the input utterance. Finally, another 1-layer CNN encoder denoted by $enc_{shared}$ is used to learn shared representations across languages. 

Given utterance $u$ as input, the final hidden states of the tokens $h_1\ldots h_T$ produced by $enc_{bert}$ are used as the inputs of the CNN encoders. Each CNN encoder, $enc$, generates its token and sentence representations, namely $h^{enc}_t$ and $s^{enc}_u$:
\begin{equation}
\begin{split}
h^{enc}_1, \ldots, h^{enc}_T = \text{CNN}^{enc}(h_1,\ldots,h_T) \\
s^{enc}_u = \text{max\_pooling}(h^{enc}_1,\ldots ,h^{enc}_T)
\end{split}
\label{eq:pri}
\end{equation}

In addition to the normal IC and SF decoders, we add two more decoders: a language predictor and a language discriminator. Each of the additional decoders consists of a feed-forward of 1 dense layer ($\text{FFN}$) and a linear output layer to predict the language of $u$. 

The language predictor receives $s^{enc_{lang}}_u$, which is the sentence representation computed by $enc_{lang}$, as input, and outputs a language distribution $p^P$:
\begin{equation}
\begin{split}
\hat{s}^{P}_u  =  \text{FFN}^P(s^{enc_{lang}}_u) \\
p^P(.|u)  =  \text{softmax}({\bf W}^P\hat{s}^{P}_u + b^P)
\end{split}
\end{equation}

The language-specific token and sentence representations are computed as following:
\begin{equation}
\begin{split}
h^{specific}_t  = \sum_{l \in {l_1, \ldots , l_N}} h^{enc_l}_t * p^P(l|u) \\
s^{specific}_u  = \sum_{l \in {l_1, \ldots , l_N}} s^{enc_l}_u * p^P(l|u) 
\end{split}
\end{equation}

The combined token and sentence representations are computed as following:
\begin{equation}
\begin{split}
h^{combined}_t  =  h^{specific}_t \oplus h^{enc_{shared}}_t  \\
s^{combined}_u  =  s^{specific}_u  \oplus s^{enc_{shared}}_u
\end{split}
\label{eq:com}
\end{equation}

The combined token and sentence representations are used as inputs to the IC and SF decoders as described in Sec. \ref{slu-bert}.

The language discriminator receives the sentence representation from $enc_{shared}$ as input, and also predicts the language distribution of $u$ as the language predictor:
\begin{equation}
\begin{split}
\hat{s}^{D}_u  =  \text{FFN}^D(s^{enc_{shared}}_u) \\
p^D(.|u)  =  \text{softmax}({\bf W}^D\hat{s}^{D}_u + b^D)
\end{split}
\end{equation}
However, the discriminator is trained adversarially to confuse the system about the distinction between languages. This encourages the model to learn shared-language representations.

We use CRF loss for the SF task, denoted by $L_s$. Meanwhile, for the IC decoder, language predictor, and language discriminator, we use cross-entropy losses denoted by  $L_i, L_p$ and $L_d$, respectively. The full model is trained by an adversarial training strategy as shown in Alg. \ref{alg:adv}. 

\begin{algorithm}
\SetAlgoLined
\KwInput{SLU training data in multiple languages with intent, slot and language labels. $\alpha_d, \alpha_i, \alpha_s, \alpha_p, \beta_d$ are model hyper-parameters.}
Task 1: To optimize $L_1 = \alpha_d L_d$ \\
Task 2: To optimize $L_2 = \alpha_i L_i + \alpha_s L_s + \alpha_p L_p - \beta_d L_d$   \\
 \While{number\_of\_epochs(Task 1) $<$ K or  number\_of\_epochs(Task 2) $<$ K}{
    Randomly pick a task $j$ ($j \in \{1, 2\}$) \\
        Generates a mini-batch for Task j  \\
        Updates model weights to optimize $L_j$
 }
 \caption{Adversarial training strategy}
   \label{alg:adv}
\end{algorithm}

\section{An ideal multilingual SLU performance vs. a naive multilingual SLU performance}
\label{lbl:ideal-flawed}
To address the question of to what degree knowledge can be transferred across languages in multilingual SLU, in this section, we define an ideal and a naive multilingual performance. Let us consider a multilingual SLU model $\mathcal{M}$ for $n$ languages $l_1, l_2, \ldots l_n$. 
 We use the following notations:
\begin{itemize}
    \item ${\bf D}$: a training set, in which each annotated utterance can be in one of the $n$ target languages.
    \item ${\bf D}^l$: The monolingual version of $\bf D$ in language $l$. That means ${\bf D}^l$ and ${\bf D}$ contain the same utterances, but all of the utterances in ${\bf D
}^l$  are in $l$ only.
    \item ${\bf d}^l$: A subset of $\bf D$ containing all utterances which are in language $l$.
    \item $\mathcal{M(\bf T)}$: Model $\mathcal{M}$ trained on dataset $\bf T$.
    \item $eval(\mathcal{M}, l)$: An evaluation metric  evaluating the performance of $\mathcal{M}$ on the test set of language $l$.
    \item $\mathcal{M'}$: A monolingual SLU model.
\end{itemize}
Given ${\bf D}$, the average performance of $\mathcal{M}$ on the target languages is computed in Eq. \ref{eq:mavg}.
\begin{equation}
\text{avg}_\mathcal{M} = \frac{1}{n}*\sum_{l \in  l_1, l_2, \ldots l_n}eval(\mathcal{M}({\bf D}), l)
\label{eq:mavg}
\end{equation}
\subsection{Ideal case}
Ideally, the language borders are completely blurred and knowledge can be transferred freely across different languages. Given ${\bf D}$,  for each target language $l$, the multilingual performance ($eval(\mathcal{M({\bf D})}, l)$) should \textit{reach or outperform} the performance of a monolingual model trained on a similar amount of knowledge ($eval(\mathcal{M'}({{\bf D}^l}), l)$). For example, let us consider three models trained on the \textit{same} amount of knowledge (represented as training data): i) A multilingual model $xy$ which has 50\% of the training data in language $X$, and the other 50\% in language $Y$; ii) A monolingual model $x$ which has 100\% of the training data in language $X$; iii) A monolingual model $y$ which has 100\% of the training data in language $Y$. In an \textit{ideal} case, model $xy$ should have similar performance on languages $X$ and $Y$ as the respective monolingual models $x$ and $y$. 

For the rest of this paper, we will refer to $eval(\mathcal{M'}({{\bf D}^l}), l)$ as the Ideal baseline on language $l$ given $\bf D$. We also define an Ideal baseline for the average performance on the target languages in Eq. \ref{eq:mideal}. Comparing $avg_\mathcal{M}$ with $avg_{ideal}$ will indicate how close $\mathcal{M}$ is to the ideal performance.

\begin{equation}
\text{avg}_{ideal} = \frac{1}{n}*\sum_{l \in  l_1, l_2, \ldots l_n} eval(\mathcal{M'}({{\bf D}^l}), l)
\label{eq:mideal}
\end{equation}
\subsection{Naive case}
Given ${\bf D}$, in a naive multilingual case, for each target language $l$, the multilingual performance   ($eval(\mathcal{M({\bf D})}, l)$) \textit{cannot outperform} the monolingual model trained on the subset of its training data containing utterances which are in $l$ ($eval(\mathcal{M'}({{\bf d}^l}), l)$). That means, the model performance in a language can not be improved by adding more training utterances from other languages. Let us reconsider the previous example. In a naive case, in which the language border between $X$ and $Y$ is completely closed, the performance of model $xy$ on  languages $X$ and $Y$ should be worse than the respective monolingual models $x$ and $y$. 

For the rest of this paper, we will refer to $eval(\mathcal{M'}({{\bf d}^l}), l)$ as the Naive baseline on language $l$ given $\bf D$. A Naive baseline for the average performance on the target languages can be found in Eq. \ref{eq:mflawed}. Comparing $avg_\mathcal{M}$ with $avg_{naive}$ will indicate how far $\mathcal{M}$ is from being naive.

\begin{equation}
\text{avg}_{naive} = \frac{1}{n}*\sum_{l \in  l_1, l_2, \ldots l_n} eval(\mathcal{M'}({{\bf d}^l}), l) 
\label{eq:mflawed}
\end{equation}

\section{Datasets}
We evaluate our model on a resampled version of the publicly available MultiATIS++ dataset \cite{multiatis++}, which contains parallel SLU data from several languages, and thus allows evaluating performance in relation to the ideal and naive baselines. In addition, we evaluate our model on real-world SLU data to explore the impact of our proposed adversarial model architecture in a real-world scenario.
\subsection{Publicly available data}
The ATIS (``Air Travel Information Service") dataset \cite{atis} is one of the most well-known datasets for evaluating SLU models. It was created by having participants solve given air travel planning scenarios, and then the resulting queries were manually transcribed and annotated. MultiATIS \cite{multiatis} extended the original dataset by adding two additional languages (Turkish and Hindi), and recently six further languages (Spanish, German, Chinese, Japanese, Portuguese, French) were added yielding MultiATIS++ \cite{multiatis++}. For our experiments, we focus on four languages for which we have both real-world and MultiATIS++ data available, i.e. English (EN), German (DE), Spanish (ES) and Japanese (JA).

In MultiATIS++, the data for additional languages were created by manually translating the original English utterances. For translations, several of the slot values, such as city names, were simply kept, yielding e.g. German requests about American airports. We consider keeping a large numbers of slot values the same across languages as being unrealistic, as in real-world applications typically slot value usage differs across locales. In addition, this makes it unrealistically easy for evaluating multilingual models, which may show strong performance just by remembering slot values from the source language.    
To make the data more realistic we resampled slot values for the city name slot, which is the most frequent one in the data. In particular, we replaced slot values for the city name slot using the cities with the biggest airports in the region, i.e. the top 30 European airport cities for DE, the top 24 Latin-American airport cities for ES and the top 28 Japanese airport cities for JA.

Across the four languages considered for our experiments, MultiATIS++ (resampled) comprises 4488, 490 and 893 utterances for train, dev and test respectively. The data cover 18 intents and 84 slots. 

\paragraph{Balanced and Imbalanced multilingual datasets} We created two variants of the resampled data to allow detailed evaluation of our proposed approach in relation to different \textit{multilingual} SLU scenarios, i.e. a balanced and an imbalanced variant. The imbalanced version reflects a real-world scenario, where usually an application is rolled out to different locales over time, yielding different data amounts, with the English version usually being the first and having highest data amounts. 

To create trilingual balanced versions, for each train and dev, we split the utterance ids\footnote{Each utterance id refers to four translated utterances in EN, DE, ES and JA.} into three equal parts, and we construct mixed datasets by taking disjoint subsets across languages. To create imbalanced versions, the train and dev data were split into three parts across languages: 50\%, 33\%, and 17\%. The largest subset corresponds to EN data, the second largest to DE data, and the smallest part is either ES or JA data for the two sets of three languages respectively. The goal was to select the data in a way that with multilingual data we have all the available utterances but in three languages. The split for the balanced versions was 50/50, and for the imbalanced version it was 70/30, with 70\% for EN and 30\% for DE or JA.

Each of the target languages has a test set taken from MultiATIS++ (resampled).
\subsection{Real-world data}
We created a dataset comprising real-world data by extracting random samples from a commercial large-scale SLU system. The data are representative of user requests to voice-controlled devices, and they were manually annotated with slot and intent labels. Aiming to get a diverse dataset, we included data from three domains\footnote{In real-world SLU applications usually the intents are split into different domains; the data in our experiments are still all from the same overall source.} (\emph{Music}, \emph{Books}, \emph{Video}). To reflect the real-world use case, where user frequency and hence data size differs across domains, we extracted samples of different sizes. In particular, we use 20k, 10k and 5k user utterances for \emph{Music}, \emph{Video} and \emph{Books}, respectively, yielding a total of 35k utterances for each language. Each domain sample was split into 80\% training, 10\% dev and 10\% test data. Note that unlike the MultiATIS++ data, the real-world data are not parallel, as they were not artificially ported from one language to another. 

\label{data}

\section{Experiments}
In this section, we first describe the multilingual performance evaluation and experimental settings. Afterwards, we present experiments on the resampled MultiATIS++ dataset and subsequently experiments with real-world-data. 
\subsection{Multilingual performance evaluation}
To address the question of to what degree language borders can be blurred in BERT-based multilingual SLU, we compare two multilingual SLU models as discussed in Sec. \ref{lbl:method} with the Ideal and Naive baselines on the Balanced and Imbalanced multilingual datasets (see Sec. \ref{data}). We denote the standard and adversarial multilingual SLU models as Multi-lang. and Lang.-adv., respectively. 
The Ideal and Naive baselines on each of the target languages are computed as in Sec. \ref{lbl:ideal-flawed} by using the BERT-based model described in Sec. \ref{slu-bert} as the monolingual model $\mathcal{M'}$.


\subsubsection{Settings} In all of our experiments, we use pre-trained multilingual BERT \cite{mBERT}. Notably, we compared pre-trained multilingual BERT and monolingual BERT in monolingual models, and found that they have similar performance on our datasets. The IC and SF decoders each have two dense layers of size 768 with gelu activation.  The dropout values used in IC and SF decoders are 0.5 and 0.2, respectively. The language decoders each have 1 dense layer of size 768 with gelu activation, and dropout value of 0.5. The CNN encoders each have one 2D convolutional layer with kernel size and hidden dim set to 3 and 512, respectively. All encoders use max-pooling for computing sentence representations.  For optimization, we use Adam optimizer with learning rate 0.1 and a Noam learning rate scheduler, and we trained our models with mini-batch size of 32 on a single GPU. For adversarial models, the $\alpha_d, \alpha_i, \alpha_s, \alpha_p, \beta_d$ hyper-parameters are set to (1.0, 1.0, 1.0, 1.0, 0.2) and (1.0, 1.0, 0.5, 0.5, 0.2) in our trilingual and bilingual experiments, respectively. The models are trained for 180 epochs with early stopping. We choose the best epoch based on the validation score of Task 2 in Alg. \ref{alg:adv}. 

To evaluate our models, we use the standard SLU metrics,  i.e.  F1 for slot filling (computed using the CoNLL2002 script) and accuracy for intent classification. In addition, following 
\newcite{gaspers2018} we use a semantic error rate, which measures IC and SF jointly and is defined as follows:
\begin{equation}
\textrm{\emph{SemER}} = \frac{\#(\textrm{slot+intent errors})}{\#\textrm{slots in reference + 1}}\end{equation}
\label{experiments}
\subsubsection{Results}
\begin{table}
\parbox{.5\linewidth}{
\centering
\resizebox{.49\textwidth}{!}{
\begin{tabular}{|l|l|l|l|l|l|l|l|l|}
\hline
\multirow{2}{*}{\textbf{Exp.}} & \multirow{2}{*}{\textbf{Target}} &  \multirow{2}{*}{\textbf{Model}}& \multicolumn{3}{|c|}{\textbf{Balanced}} & \multicolumn{3}{|c|}{\textbf{Imbalanced}} \\
\cline{4-9} 
&&&SemER&SF F1&IC acc.&SemER&SF F1&IC acc.\\
\hline
\hline
\multirow{16}{*}{\rotatebox[origin=c]{90}{\textbf{EN-DE-ES}}}&\multirow{4}{*}{EN} &Naive &  16.06 &90.12& 89.25 & 15.54  & 91.07  & 87.12  \\
& & Ideal & 6.97 & 95.53 & 97.54 & 6.97 & 95.53 & 97.54 \\
& & Multi-lang.& 8.14 & 94.68 & 96.53 & 7.46 & 95.13 & 96.53 \\
& & Lang.-adv.& 8.36 & 94.54 & 96.41 & 7.90  & 94.56 & 97.20 \\
\cline{2-9} 
&\multirow{4}{*}{DE} &Naive & 16.74  & 89.97 & 86.10 & 17.4  & 89.88  & 86.21\\
& & Ideal &8.48  & 94.47 & 95.74 & 8.48  & 94.47 & 95.74 \\
& & Multi-lang.& 9.38 & 93.97  & 95.07 & 10.77 & 93.11 & 94.51 \\
& & Lang.-adv.& 8.51 &94.69  & 95.85 & 10.17& 93.40 & 96.19 \\
\cline{2-9} 
&\multirow{4}{*}{ES} &Naive & 27.17  & 80.26  & 86.10 & 34.38& 74.09& 83.39\\
& & Ideal & 14.45 & 88.37  & 95.82 & 14.45 & 88.37  & 95.82 \\
& & Multi-lang.& 15.79 & 86.94 & 95.33 & 17.24 & 85.91 & 95.94 \\
& & Lang.-adv.& 14.14 & 88.30  & 96.68 & 16.55 & 86.19 & 95.82 \\
\cline{2-9} 
&\multirow{4}{*}{Avg.} &Naive & 19.99  & 86.78 & 87.15 &22.44 & 85.01&85.57 \\
& & Ideal & 9.97 & 92.79 & 96.37 &9.97 & 92.79 & 96.37\\
& & Multi-lang.& 11.10  & 91.86  & 95.64 & 11.82 & \textbf{91.38} & 95.73 \\
& & Lang.-adv.& \textbf{10.34} & \textbf{92.51}  & \textbf{96.31}  & \textbf{11.54 }& \textbf{91.38} & \textbf{96.40} \\
\cline{2-9} 
\hline
\hline

\end{tabular}}
}
\hfill
\parbox{.5\linewidth}{
\centering
\resizebox{.49\textwidth}{!}{
\begin{tabular}{|l|l|l|l|l|l|l|l|l|}
\hline
\multirow{2}{*}{\textbf{Exp.}} & \multirow{2}{*}{\textbf{Target}} &  \multirow{2}{*}{\textbf{Method}}& \multicolumn{3}{|c|}{\textbf{Balanced}} & \multicolumn{3}{|c|}{\textbf{Imbalanced}} \\
\cline{4-9} 
&&&SemER&SF F1&IC acc.&SemER&SF F1&IC acc.\\
\hline
\hline
\multirow{16}{*}{\rotatebox[origin=c]{90}{\textbf{EN-DE-JA}}}&\multirow{4}{*}{EN} 
&Naive & 17.25 & 89.4  & 86.9 & 10.55 & 92.89 & 95.86 \\
& & Ideal & 6.97 & 95.53 & 97.54 & 6.97 & 95.53 & 97.54  \\
& & Multi-lang. & 8.66 & 94.23  & 96.53  & 9.31 & 93.98  & 95.63\\
& & Lang.-adv.& 7.92 & 94.85 & 95.52 & 7.57  & 94.90  & 94.42\\
\cline{2-9} 
&\multirow{4}{*}{DE} &Naive & 18.32 & 88.84  & 84.87 & 17.4 & 90.10 & 85.43 \\
& & Ideal & 8.48  & 94.47 & 95.74 & 8.48  & 94.47 & 95.74 \\
& & Multi-lang.& 8.70 & 94.44  & 95.63 & 11.21 & 92.91 & 94.28 \\
& & Lang.-adv.& 8.62 & 94.62  & 95.29 & 9.77 & 94.10 & 94.62 \\
\cline{2-9} 
&\multirow{4}{*}{JA} &Naive & 16.54 & 90.70  & 87.36 & 27.1 & 84.4 & 79.91 \\
& & Ideal & 13.15 & 92.40  & 92.21  & 13.15 & 92.40  & 92.21 \\
& & Multi-lang.& 13.98 & 92.01  & 90.07 & 16.16 & 90.6 & 88.6 \\
& & Lang.-adv.& 13.82 & 92.25 & 89.28 & 16.37& 91.19 & 89.61 \\
\cline{2-9} 
&\multirow{4}{*}{Avg.} &Naive & 17.37  & 89.64 & 86.38 & 18.35&89.13 & 87.07 \\
& & Ideal& 9.53  & 94.13  & 95.16 & 9.53  & 94.13  & 95.16\\
& & Multi-lang.&   10.45 & 93.56 & \textbf{94.08} & 12.23 & 92.50& 92.84\\
& & Lang.-adv.& \textbf{10.12} & \textbf{93.91} & 93.36 &  \textbf{11.24} &\textbf{93.40} & \textbf{93.88} \\
\cline{2-9} 
\hline
\hline

\end{tabular}}
}
\caption{SLU evaluations on trilingual MultiATIS++ (resampled) datasets. Lower numbers indicate better performance for SemER, while higher numbers indicate better performance for slot F1 and intent classification accuracy. Bold numbers indicate better performance between the two multilingual SLU models.}
\label{tbl:atis_tri}
\end{table}

\begin{table}
\parbox{.5\linewidth}{
\centering
\resizebox{.49\textwidth}{!}{
\begin{tabular}{|l|l|l|l|l|l|l|l|l|}
\hline
\multirow{2}{*}{\textbf{Exp.}} & \multirow{2}{*}{\textbf{Target}} &  \multirow{2}{*}{\textbf{Model}}& \multicolumn{3}{|c|}{\textbf{Balanced}} & \multicolumn{3}{|c|}{\textbf{Imbalanced}} \\
\cline{4-9} 
&&&SemER&SF F1&IC acc.&SemER&SF F1&IC acc.\\
\hline
\hline
\multirow{12}{*}{\rotatebox[origin=c]{90}{\textbf{EN-DE}}}&\multirow{4}{*}{EN} &Naive & 9.12 & 94.12     & 95.63 & 8.22  & 94.68  & 96.98  \\
& & Ideal & 6.97 & 95.53 & 97.54 & 6.97 & 95.53 & 97.54 \\
& & Multi-lang.& 9.34& 94.06 & 94.74   & 7.24 & 95.36  & 97.31    \\
& & Lang.-adv.& 7.82 & 94.77   & 95.74 & 7.08 & 95.21  & 97.42  \\
\cline{2-9} 
&\multirow{4}{*}{DE} &Naive& 14.65 & 91.23   & 88.12 & 15.35 & 90.88  & 87.56  \\ 
& & Ideal &8.48  & 94.47 & 95.74 & 8.48  & 94.47 & 95.74 \\
& & Multi-lang.&  9.57& 93.98   & 94.96 &  8.26 & 94.17  & 96.08  \\
& & Lang.-adv.& 9.11 & 94.16   & 95.63 & 9.55  & 93.83  & 95.52   \\

\cline{2-9} 
&\multirow{4}{*}{Avg.} &Naive&  11.89 & 92.68  & 91.88 &  11.79 & 92.78  & 92.27 \\
& & Ideal & 7.73 & 95.00 & 96.64 &  7.73 & 95.00 & 96.64 \\
& & Multi-lang.& 9.46  & 93.98  & 94.85 & 8.36 & \textbf{94.77}  & 96.17  \\
& & Lang.-adv.& \textbf{8.47} &  \textbf{94.47} & \textbf{95.69} & \textbf{8.32} & 94.52  & \textbf{96.47}  \\
\cline{2-9} 
\hline
\hline

\end{tabular}}
}
\hfill
\parbox{.5\linewidth}{
\centering
\resizebox{.49\textwidth}{!}{
\begin{tabular}{|l|l|l|l|l|l|l|l|l|}
\hline
\multirow{2}{*}{\textbf{Exp.}} & \multirow{2}{*}{\textbf{Target}} &  \multirow{2}{*}{\textbf{Method}}& \multicolumn{3}{|c|}{\textbf{Balanced}} & \multicolumn{3}{|c|}{\textbf{Imbalanced}} \\
\cline{4-9} 
&&&SemER&SF F1&IC acc.&SemER&SF F1&IC acc.\\
\hline
\hline
\multirow{12}{*}{\rotatebox[origin=c]{90}{\textbf{EN-JA}}}&\multirow{4}{*}{EN} 
&Naive & 10.29 & 93.03  & 95.97 & 8.96 & 94.19  & 94.62  \\
& & Ideal & 6.97 & 95.53 & 97.54 & 6.97 & 95.53 & 97.54  \\
& & Multi-lang. & 8.87 & 94.01   & 96.97   & 8.82 & 94.39   & 95.63\\
& & Lang.-adv.& 8.55 & 94.28  & 95.86  & 7.54  & 95.11   & 96.86 \\
\cline{2-9} 
&\multirow{4}{*}{JA} &Naive & 18.13 & 89.71   & 85.21 & 20.77  & 87.94  & 84.09  \\
& & Ideal & 13.15 & 92.40  & 92.21  & 13.15 & 92.40  & 92.21 \\
& & Multi-lang. & 13.87 &  92.35  & 89.28   & 15.44& 91.71   & 86.34\\
& & Lang.-adv.& 13.21 & 92.81  & 89.05  & 14.37  & 92.41   & 87.70 \\
\cline{2-9} 
&\multirow{4}{*}{Avg.} &Naive &14.21  & 91.37  & 90.59 & 15.32  & 91.07  & 89.36 \\
& & Ideal& 10.06 &  93.97 &94.88 &10.06 &  93.97 &94.88  \\
& & Multi-lang. & 11.37& 93.18  & \textbf{93.13}   & 12.13 & 93.05   & 90.99\\
& & Lang.-adv.& \textbf{10.88} & \textbf{93.55} & 92.46  & \textbf{10.96}  & \textbf{93.76 }  & \textbf{92.28} \\
\cline{2-9} 

\hline
\hline

\end{tabular}}
}
\caption{SLU evaluations on bilingual MultiATIS++ (resampled) datasets. Lower numbers indicate better performance for SemER, while higher numbers indicate better performance for slot F1 and intent classification accuracy. Bold numbers indicate better performance between the two multilingual SLU models.}
\label{tbl:atis_bi}
\end{table}
\begin{figure}[ht]
    \parbox{0.5\linewidth}{
        \centering
        \resizebox{0.5\textwidth}{!}{
            \centering
            \includegraphics[width=9cm]{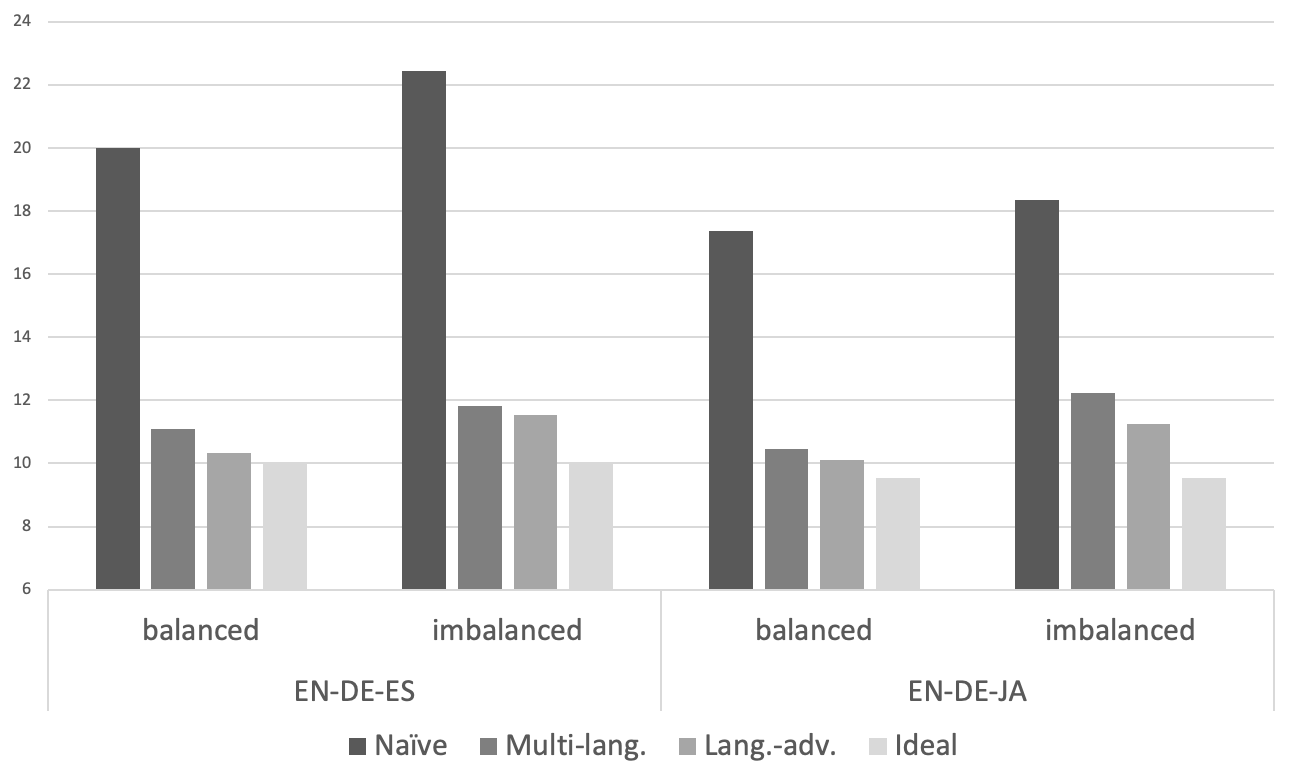}
        }
    }
    \parbox{0.5\linewidth}{
        \centering
        \resizebox{0.5\textwidth}{!}{
            \centering
            \includegraphics[width=9cm]{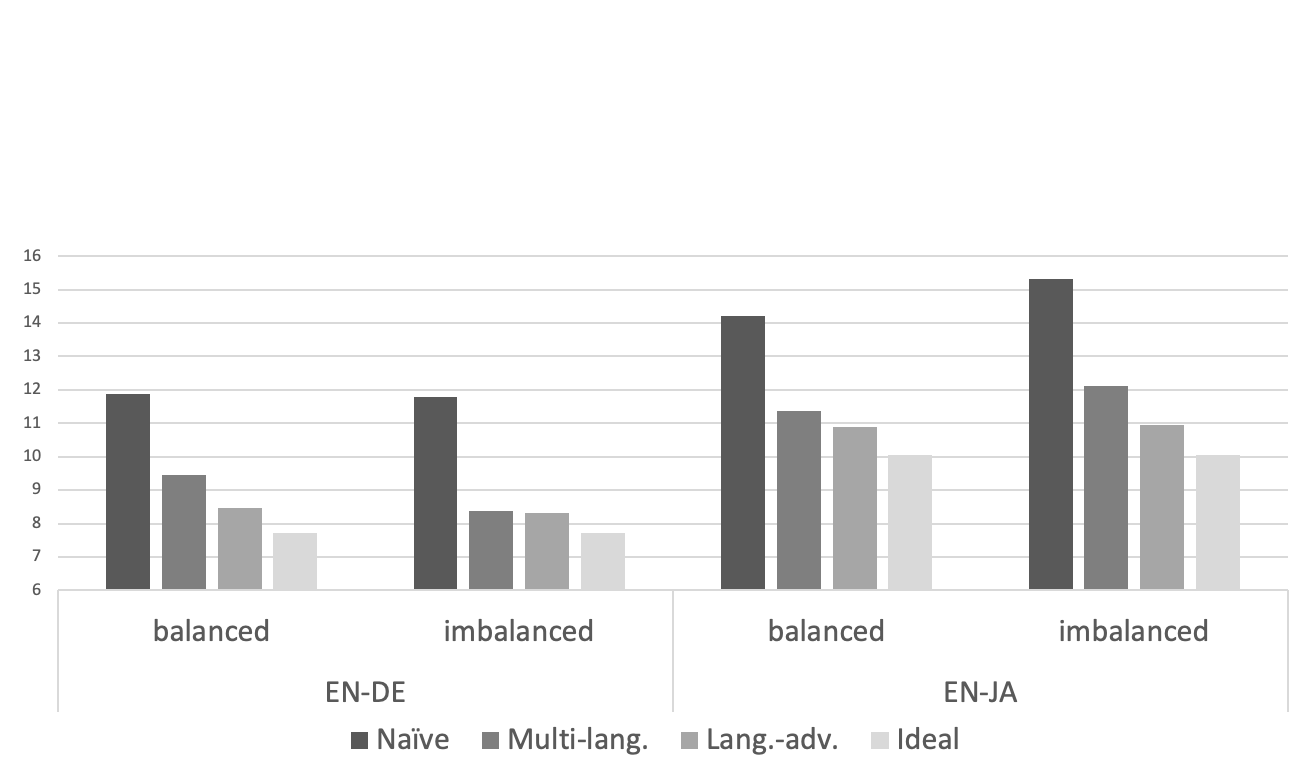}
        }
    }
    \caption{Average semantic error rate (SemER) comparisons on MultiATIS++ (resampled).}
    \label{fig:semer_compare}
\end{figure}

Table \ref{tbl:atis_bi} and Table \ref{tbl:atis_tri} show our experimental results on bilingual and trilingual datasets, respectively. As can be seen, both of the two multilingual SLU models consistently outperform the Naive baseline, meaning that multilingual techniques can effectively transfer knowledge across different languages. The relative improvements in avg. SemER range from 19.98\% on Balanced (EN-JA) to  48.58\% on Imbalanced (EN-DE-ES) scenarios. Interestingly, this holds even on the groups of (EN, JA) and (EN, DE, JA) in which JA is linguistically distant to the other languages. Moreover, the multilingual models surpass Naive performance not only on the average performance over the target languages but also on every individual language. 

Overall, there is still a gap between the Ideal performance and the two multilingual SLU models. The largest difference in avg. SemER is observed on Imbalanced(EN-DE-JA) where Multi-lang. is 28.33\% relatively worse than Ideal. However, the fact that the absolute gap of 2.7\% here is rather small compared to the absolute improvement of 6.12\%  over Naive baseline, confirms the effectiveness of the multilingual techniques. 


The experimental results also prove our hypothesis that using both language-shared and language-specific features can narrow the gap to the Ideal performance. In particular, Lang.-adv. gives lower average semantic error rates (SemER) than Multi-lang. in all of our scenarios. The other two metrics, SF F1 and IC acc., are also improved by Lang.-adv. over  Multi-lang. in 5 out of 6 datasets. Surprisingly,  Lang.-adv. even exceeds or comes very close to the Ideal performance for ES and DE on the dataset Balanced (EN-DE-ES), and for DE on the dataset Balanced (EN, DE, JA). In addition, the results show that there is a significant difference in model performance on the balanced and imbalanced versions of the datasets. Across the different trilingual and bilingual settings, the imbalanced versions of the dataset have on average a lower performance than their balanced counterparts. This shows that while transferring knowledge across different languages works, there is still a drop in performance if data available for one of the languages is scarce.

Another interesting finding can be seen in the comparison of the two bilingual settings, as the language pair EN-DE consistently performs better than EN-JA. This is to be expected as JA is more distant to EN than EN is to DE as EN and DE are both Germanic languages.

Fig. \ref{fig:semer_compare} visualizes the comparisons of the two multilingual SLU models against the Naive and Ideal performance. 

\subsubsection{Comparison to state-of-the-art and model complexity}
Using the same training, development and testing data as \newcite{multiatis++}, our monolingual SLU model obtains 97.54 intent accuracy and 95.53 slot F1 on English ATIS (Ideal baseline on English). These results are comparable to the BERT-based performances reported by \newcite{multiatis++}, which are 97.20 intent accuracy and 95.57 slot F1. This confirms the strength of our baselines. 

A common concern with a BERT-based model is about its complexity. In fact, using CNN encoders prevents our adversarial model Lang.-adv. from a dramatic increase in the number of parameters. On the trilingual datasets, the number of parameters is about 190.4M which is just 5.6\% more than that of Multi-lang. We also found that replacing CNN encoders by BERT encoders not only increases the number of model parameters to 260.9M but also does not bring significant gain in model performance. 
\subsection{Performance on real-world SLU data}
We investigated performance of our SLU models on real-world data to explore the impact on a real-world scenario. Since parallel data were not available for these experiments, we cannot provide performance in relation to an Ideal baseline. Instead, we use the Naive baseline and determine the relative gain obtained by the two multilingual SLU models. The results for two language triples (EN, DE, ES) and (EN, DE, JA) are presented in Table \ref{results_alexa}.
\begin{table}[ht]
\centering
\resizebox{0.8 \textwidth}{!} {
\begin{tabular}{|l|l|l|l|l|l|l|l|l|l|l|l|} 
\hline
\multirow{2}{*}{\textbf{Exp.}} & \multirow{2}{*}{\textbf{Target}} &  \multirow{2}{*}{\textbf{Model}}& \multicolumn{3}{|c|}{\textbf{Music domain}} & \multicolumn{3}{|c|}{\textbf{Books domain}} & \multicolumn{3}{|c|}{\textbf{Video domain}} \\
\cline{4-12} 
&&&$\Delta$ SemER&$\Delta$ SF F1&$\Delta$ IC acc.&$\Delta$ SemER&$\Delta$ SF F1&$\Delta$ IC acc.&$\Delta$ SemER&$\Delta$ SF F1&$\Delta$ IC acc.\\
\hline
\hline
\multirow{8}{*}{\rotatebox[origin=c]{90}{\textbf{EN-DE-ES}}}
&\multirow{2}{*}{EN} &Multi-lang.&-24.14 &+7.18 &+3.07 & -16.53&+2.64 &+3.62&-10.41 &+3.18 &+3.98\\
& & Lang.-adv.&-24.81&+7.03 & +3.34& -29.84&+4.58 & +7.69&-18.8&+6.21 &+4.77 \\
\cline{2-12} 
&\multirow{2}{*}{DE} &Multi-lang.&-20.06 & +5.24& +3.19& -9.2 &+0.68 &+5.62&-16.00 &+6.6 &+3.48\\
& & Lang.-adv.&-23.89& +6.1& +3.46&-28.02 &+3.23 &+11.24&-25.22 &+10.48 &+4.46\\
\cline{2-12} 
&\multirow{2}{*}{ES} &Multi-lang.&-27.85 &+9.29 &+3.13 & -18.25&+2.1 &+8.45&-15.5 &+5.41 &+4.97\\
& & Lang.-adv.&-29.18& +9.6&+3.18 &-29.17 &+4.63 &+12.44 &-20.44& +7.81&+5.97\\
\cline{2-12} 
&\multirow{2}{*}{Avg.} &Multi-lang. & -24.02 & +7.24 & +3.13 &  -14.66 &+1.81 &+5.89 & -13.97 & +4.92 &+4.14 \\
& & Lang.-adv.& \textbf{-25.96} & \textbf{+7.58 }& \textbf{+3.33} & \textbf{-29.01} & \textbf{+4.15} & \textbf{+10.46} & \textbf{-21.49} & \textbf{+8.17}&\textbf{+5.07} \\
\hline
\hline
\multirow{8}{*}{\rotatebox[origin=c]{90}{\textbf{EN-DE-JA}}}
&\multirow{2}{*}{EN} &Multi-lang.& -19.5&+5.98 & +2.86&-12.88 &+1.15 &+4.07& -5.67&+1.96 &+3.18\\
& & Lang.-adv.&-25.32&+7.37 & +3.28&-27.4 &+4.32 &+6.56 &-17.3&+5.28 & +4.77\\
\cline{2-12} 
&\multirow{2}{*}{DE} &Multi-lang.& -20.67&+5.17 &+3.08 &-8.9 &+0.53 &+3.98& -10.4& +4.4&+2.07\\
& & Lang.-adv.&-24.66& +6.67& +3.24&-23.56 & +2.7&+9.6 &-20.89&+8.14 & +4.46\\
\cline{2-12} 
&\multirow{2}{*}{JA} &Multi-lang.&-27.82 &+13.36 &+2.28 & -3.9&-3.41 &+6.49&-0.1 &-6.5 &+3.18\\
& & Lang.-adv.&-31.9& +14.49& +2.81&-13.67 &+0.77 &+12.26 &-17.93&+2.47 &+6.93 \\
\cline{2-12} 
&\multirow{2}{*}{Avg.} &Multi-lang. & -22.66 & +8.17 & +2.74 & -8.56  & -0.58 & +4.85& -5.39 & -0.05 & +2.81 \\
& & Lang.-adv.&\textbf{ -27.29} & \textbf{+9.51} & \textbf{+3.11 }&  \textbf{-21.54}& \textbf{+2.60} & \textbf{+9.47} & \textbf{-18.71} &\textbf{+5.30} & \textbf{+5.39}  \\
\hline
\hline

\end{tabular}
}
\caption{Relative change in semantic error rate (SemER), intent classification accuracy and slot F1 for multilingual and language-adversarial training compared to the Naive baseline. Negative numbers indicate better performance for SemER, while positive numbers indicate better performance for slot F1 and intent classification accuracy.} 
\label{results_alexa}
\end{table}

Overall, multilingual models improves performance over the Naive baselines. In particular, the only drops in performance are observed for slot filling in JA with Multi-lang. on two out of the three domains. This drop may be attributed to JA being linguistically distant to the other languages, which are all from the European language family. The highest relative gain is up to 27.29\% in avg. SemER obtained with Lang.-adv.

Between the two multilingual models, our proposed model, Lang.-adv. outperforms Multi-lang. in all of the three domains. While the relative gains in avg. SemER of Multi-lang. fluctuate between 5.39\% and 24.02\%, those of Lang.-adv. range between 18.71\% to 29.01\%. Moreover, unlike the standard multilingual approach, Lang.-adv. models do not yield drops when the distant language JA is included in training. In fact, their relative SemER reductions for JA range from 13.67\% to 31.9\%. Thus, in particular, when distant languages are included in multilingual training, dividing the model into language-shared and language-specific parts seems to be beneficial. Taken together, the results indicate that our proposed adversarial architecture improves performance over the standard BERT-based approach not just on academic benchmark datasets, but also on real-world SLU data.

\section{Conclusion}
We have addressed the question of to what degree language borders can be blurred in BERT-based multilingual SLU. Our experimental results on a wide range of multilingual SLU datasets showed that although mBERT is substantially good in blurring language borders even on distant language groups, there is still a gap to the ideal multilingual performance. To narrow this gap, we proposed an adversarial model architecture which uses BERT and CNN encoders to learn language-shared and language-specific representations for SLU. 

\label{conclusion}


%
%

\bibliographystyle{coling}
\bibliography{coling2020}

\begin{thebibliography}{}

\bibitem[\protect\citename{Chen \bgroup et al.\egroup }2019a]{chen19}
Q.~Chen, Z.~Zhuo, and W.~Wang.
\newblock 2019a.
\newblock Bert for joint intent classification and slot filling.
\newblock {\em arXiv:1902.10909}.

\bibitem[\protect\citename{Chen \bgroup et al.\egroup
  }2019b]{chen-etal-2019-multi-source}
Xilun Chen, Ahmed~Hassan Awadallah, Hany Hassan, Wei Wang, and Claire Cardie.
\newblock 2019b.
\newblock Multi-source cross-lingual model transfer: Learning what to share.
\newblock In {\em Proceedings of the 57th Annual Meeting of the Association for
  Computational Linguistics}, pages 3098--3112, Florence, Italy, July.
  Association for Computational Linguistics.

\bibitem[\protect\citename{Devlin \bgroup et al.\egroup }2018]{mBERT}
Jacob Devlin, Ming{-}Wei Chang, Kenton Lee, and Kristina Toutanova.
\newblock 2018.
\newblock {BERT:} pre-training of deep bidirectional transformers for language
  understanding.
\newblock {\em CoRR}, abs/1810.04805.

\bibitem[\protect\citename{Do and Gaspers}2019a]{do-gaspers-2019-cross}
Quynh Do and Judith Gaspers.
\newblock 2019a.
\newblock Cross-lingual transfer learning with data selection for large-scale
  spoken language understanding.
\newblock In {\em Proceedings of the 2019 Conference on Empirical Methods in
  Natural Language Processing and the 9th International Joint Conference on
  Natural Language Processing (EMNLP-IJCNLP)}, pages 1455--1460, Hong Kong,
  China, November. Association for Computational Linguistics.

\bibitem[\protect\citename{Do and Gaspers}2019b]{do19}
Quynh Ngoc~Thi Do and Judith Gaspers.
\newblock 2019b.
\newblock Cross-lingual transfer learning for spoken language understanding.
\newblock {\em Proceedings of the 2019 IEEE International Conference on
  Acoustics, Speech and Signal Processing, ICASSP}.

\bibitem[\protect\citename{Gaspers \bgroup et al.\egroup }2018]{gaspers2018}
Judith Gaspers, Penny Karanasou, and Rajen Chatterjee.
\newblock 2018.
\newblock Selecting machine-translated data for quick bootstrapping of a
  natural language understanding system.
\newblock {\em Proceedings of NAACL-HLT}.

\bibitem[\protect\citename{Gaspers \bgroup et al.\egroup
  }2020]{gaspers2020data}
Judith Gaspers, Quynh Do, and Fabian Triefenbach.
\newblock 2020.
\newblock Data balancing for boosting performance of low-frequency classes in
  spoken language understanding.
\newblock In {\em INTERSPEECH}.

\bibitem[\protect\citename{{He} \bgroup et al.\egroup }2020]{8990095}
K.~{He}, W.~{Xu}, and Y.~{Yan}.
\newblock 2020.
\newblock Multi-level cross-lingual transfer learning with language shared and
  specific knowledge for spoken language understanding.
\newblock {\em IEEE Access}, 8:29407--29416.

\bibitem[\protect\citename{Johnson \bgroup et al.\egroup
  }2019]{johnson-etal-2019-cross}
Andrew Johnson, Penny Karanasou, Judith Gaspers, and Dietrich Klakow.
\newblock 2019.
\newblock Cross-lingual transfer learning for {J}apanese named entity
  recognition.
\newblock In {\em Proceedings of the 2019 Conference of the North {A}merican
  Chapter of the Association for Computational Linguistics: Human Language
  Technologies, Volume 2 (Industry Papers)}, pages 182--189, Minneapolis,
  Minnesota, June. Association for Computational Linguistics.

\bibitem[\protect\citename{Liu and Lane}2016]{Liu2016AttentionBasedRN}
Bing Liu and Ian Lane.
\newblock 2016.
\newblock Attention-based recurrent neural network models for joint intent
  detection and slot filling.
\newblock In {\em INTERSPEECH}.

\bibitem[\protect\citename{Pires \bgroup et al.\egroup
  }2019]{pires2019multilingual}
Telmo Pires, Eva Schlinger, and Dan Garrette.
\newblock 2019.
\newblock How multilingual is multilingual {BERT}?
\newblock In {\em Proceedings of the 57th Annual Meeting of the Association for
  Computational Linguistics}, pages 4996--5001, Florence, Italy, July.
  Association for Computational Linguistics.

\bibitem[\protect\citename{Price}1990]{atis}
P.~J. Price.
\newblock 1990.
\newblock Evaluation of spoken language systems: The atis domain.
\newblock In {\em Proceedings of the Workshop on Speech and Natural Language},
  HLT ’90, page 91–95, USA. Association for Computational Linguistics.

\bibitem[\protect\citename{{Upadhyay} \bgroup et al.\egroup }2018]{multiatis}
S.~{Upadhyay}, M.~{Faruqui}, G.~{Tür}, H.~{Dilek}, and L.~{Heck}.
\newblock 2018.
\newblock (almost) zero-shot cross-lingual spoken language understanding.
\newblock In {\em 2018 IEEE International Conference on Acoustics, Speech and
  Signal Processing (ICASSP)}, pages 6034--6038.

\bibitem[\protect\citename{Wu and Dredze}2019]{wu-dredze-2019-beto}
Shijie Wu and Mark Dredze.
\newblock 2019.
\newblock Beto, bentz, becas: The surprising cross-lingual effectiveness of
  {BERT}.
\newblock In {\em Proceedings of the 2019 Conference on Empirical Methods in
  Natural Language Processing and the 9th International Joint Conference on
  Natural Language Processing (EMNLP-IJCNLP)}, pages 833--844, Hong Kong,
  China, November. Association for Computational Linguistics.

\bibitem[\protect\citename{Xu \bgroup et al.\egroup }2020]{multiatis++}
Weijia Xu, Batool Haider, and Saab Mansour.
\newblock 2020.
\newblock End-to-end slot alignment and recognition for cross-lingual nlu.
\newblock {\em arXiv:2004.14353}.

\bibitem[\protect\citename{Zhou and Xu}2015]{zhou-xu-2015-end}
Jie Zhou and Wei Xu.
\newblock 2015.
\newblock End-to-end learning of semantic role labeling using recurrent neural
  networks.
\newblock In {\em Proceedings of the 53rd Annual Meeting of the Association for
  Computational Linguistics and the 7th International Joint Conference on
  Natural Language Processing (Volume 1: Long Papers)}, pages 1127--1137,
  Beijing, China, July. Association for Computational Linguistics.

\end{thebibliography}

\end{document}